\documentclass[10pt,twocolumn,letterpaper]{article}

\usepackage{wacv}
\usepackage{times}
\usepackage{epsfig}
\usepackage{graphicx}
\usepackage{amsmath}
\usepackage{amsthm}
\usepackage{amssymb}
\usepackage{xcolor}
\usepackage{verbatim}
\usepackage{algorithm}
\usepackage{breqn}
\usepackage{multirow}
\usepackage{dsfont}
\usepackage{caption}
\usepackage[noend]{algpseudocode}
\graphicspath{ {./} }
\definecolor{darkspringgreen}{rgb}{0.09, 0.45, 0.27}
\usepackage[utf8]{inputenc}
\usepackage[english]{babel}
\usepackage{algorithm}
\usepackage{multirow}
\usepackage{dsfont}
\usepackage[font=Large]{subcaption}
\usepackage{caption}
\usepackage[noend]{algpseudocode}
\usepackage{epstopdf}
\usepackage{amsfonts}

\newcommand{\bI}{\mathbf{I}}

\newcommand{\bY}{\mathbf{Y}}

\newtheorem{theorem}{Theorem}
\newtheorem{corollary}{Corollary}

\wacvfinalcopy 

\def\wacvPaperID{592}

\ifwacvfinal\fi
\begin{document}

\title{Distribution-Aware Binarization of Neural Networks for Sketch Recognition}

\author{Ameya Prabhu\quad
Vishal Batchu\quad
Sri Aurobindo Munagala\quad
Rohit Gajawada\quad
Anoop Namboodiri\\
Center for Visual Information Technology, Kohli Center on Intelligent Systems\\
IIIT-Hyderabad, India\\
{\tt\small \{ameya.prabhu@research., vishal.batchu@students.,  s.munagala@research.},\\{\tt\small rohit.gajawada@students., anoop@\}iiit.ac.in}}

\maketitle
\ifwacvfinal\thispagestyle{empty}\fi

\begin{abstract}
Deep neural networks are highly effective at a range of computational tasks. However, they tend to be computationally expensive, especially in vision-related problems, and also have large memory requirements. One of the most effective methods to achieve significant improvements in computational/spatial efficiency is to binarize the weights and activations in a network. However, naive binarization results in accuracy drops when applied to networks for most tasks. In this work, we present a highly generalized, distribution-aware approach to binarizing deep networks that allows us to retain the advantages of a binarized network, while reducing  accuracy drops. We also develop efficient implementations for our proposed approach across different architectures. We present a theoretical analysis of the technique to show the effective representational power of the resulting layers, and explore the forms of data they model best. Experiments on popular datasets show that our technique offers better accuracies than naive binarization, while retaining the same benefits that binarization provides - with respect to run-time compression, reduction of computational costs, and power consumption.
\end{abstract}

\section{Introduction}


Deep learning models are pushing the state-of-the-art in various problems across domains, but are computationally intensive to train and run, especially Convolutional Neural Networks (CNNs) used for vision applications. They also occupy a large amount of memory, and the amount of computation required to train a network leads to high power consumption as well.
 
There have been many developments in the area of model compression in the last few years, with the aim of bringing down network runtimes and storage requirements to mobile-friendly levels. Compression strategies for Convolutional Neural Networks included architectural improvements \cite{he2016deep, iandola2016squeezenet} and re-parametrization \cite{moczulski2015acdc, yang2015deep} to pruning techniques \cite{han2015deep, liu2017learning} and quantization \cite{hubara2016quantized,zhou2016dorefa}. Among these approaches, quantization - especially, binarization - provided the most compact models as shown in Table \ref{table:versions_typesofcompression}.

Quantized networks - where weights/activations were quantized into low-precision representations - were found to achieve great model compression. Quantization has proven to be a powerful compression strategy, especially the most extreme form of quantization - Binarization. Binarization has enabled the use of XNOR-Popcount operations for vector dot products, which take much less time compared to full-precision Multiply-Accumulates (MACs), contributing to a huge speedup in convolutional layers \cite{rastegari2016xnor,hubara2016quantized} on a general-purpose CPU. Moreover, as each binary weight requires only a single bit to represent, one can achieve drastic reductions in run-time memory requirements. Previous research \cite{rastegari2016xnor,hubara2016quantized} shows that it is possible to perform weight and activation binarization on large networks with up to 58x speedups and approximately 32x compression ratios, albeit with significant drops in accuracy. 
\begin{figure}[t]
           \hspace{-0.6cm}\includegraphics[width=0.55\textwidth]{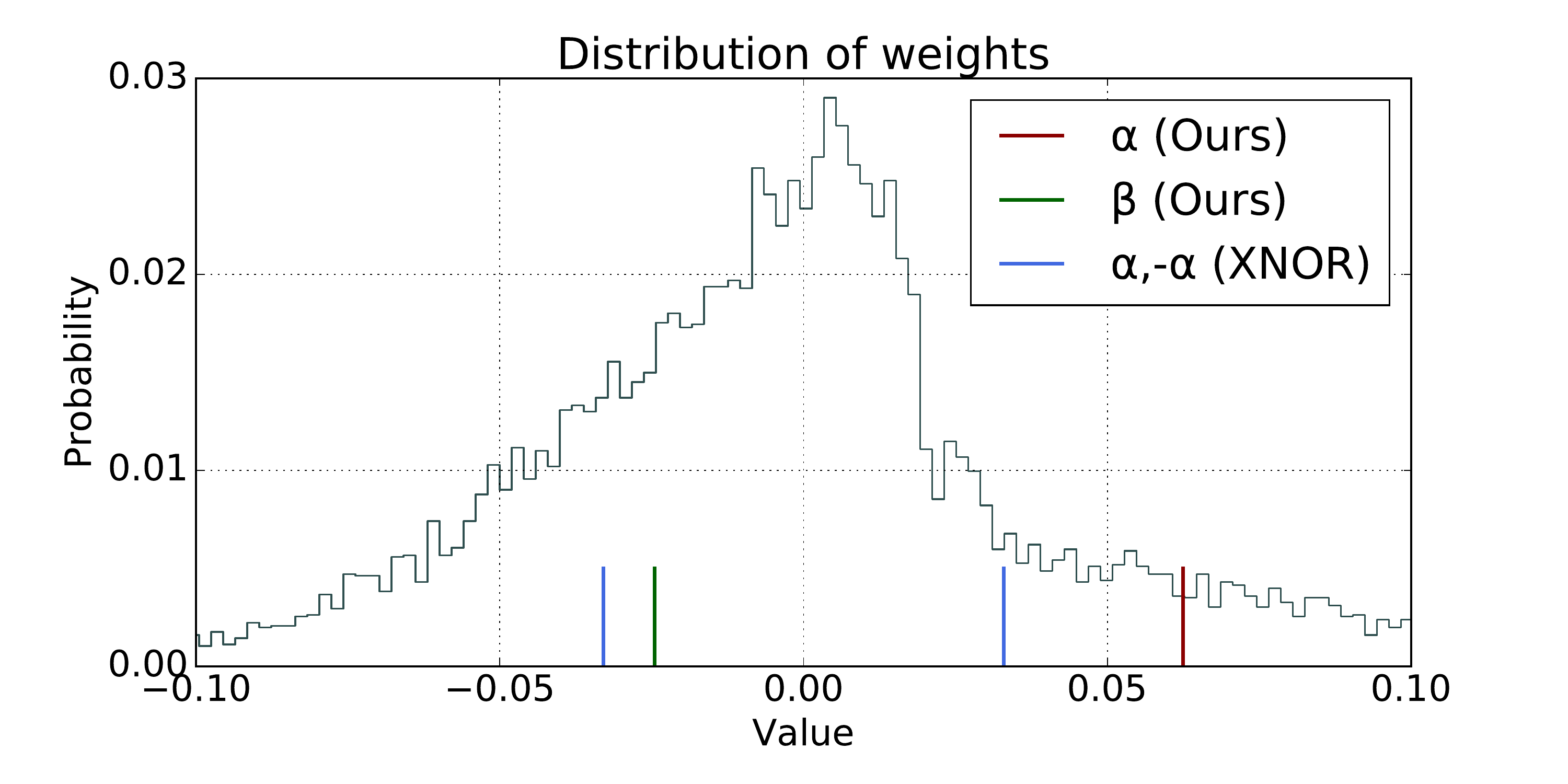}
           \caption{
           Weight distribution of a layer with corresponding $\alpha$/$\beta$ values, and the scaling factor $\alpha$ in the XNOR-Net implementation for comparison. $\alpha$ and $\beta$ in our method have differing magnitudes, unlike in XNOR-Net.}
        \label{fig:weightdistributionsample}
        \vspace{-0.4cm}
\end{figure}
\begin{figure}[t]
           \includegraphics[width=0.45\textwidth]{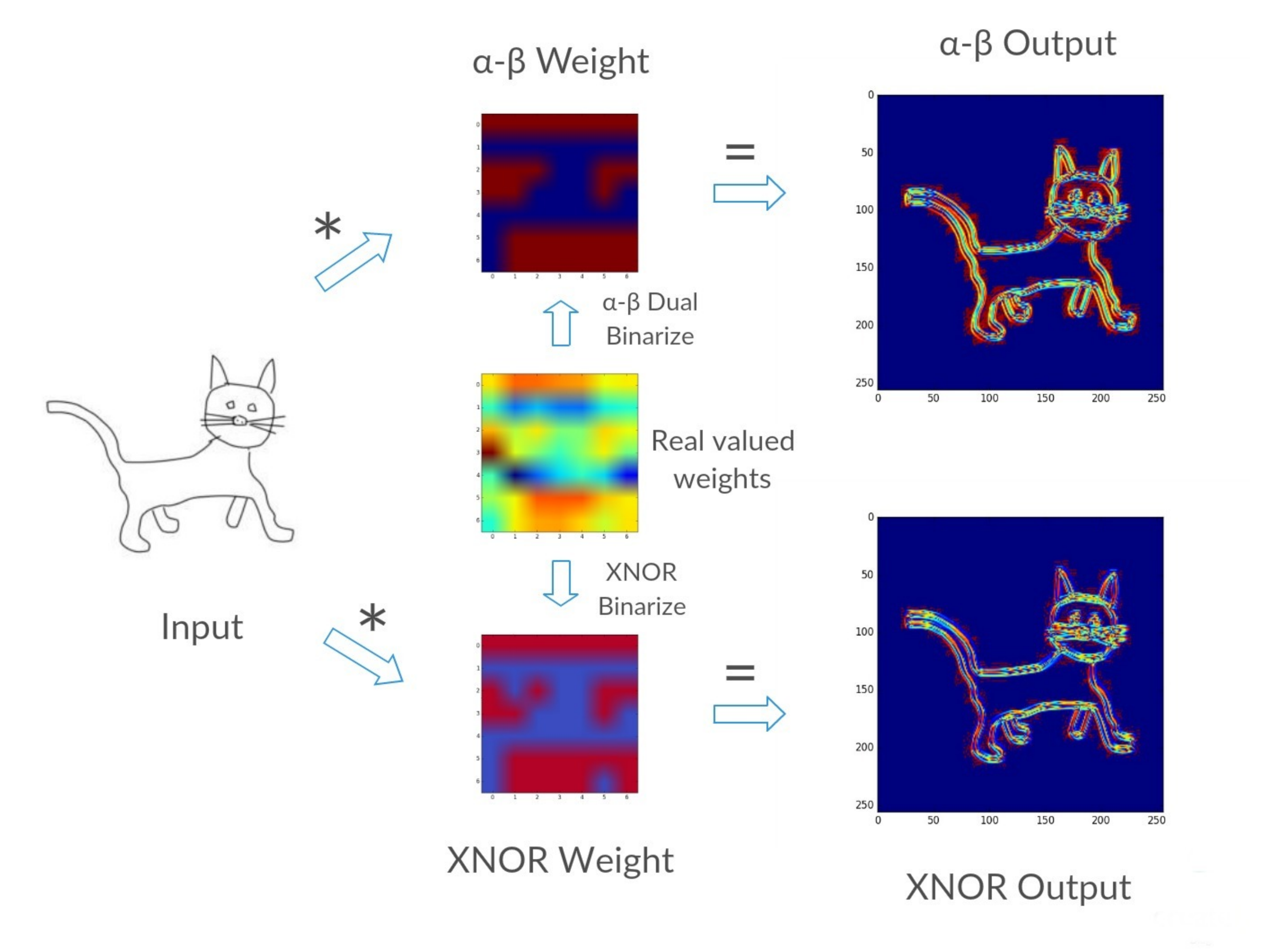}
           \caption{An example sketch passing through a convolutional layer filter, with the real-valued filter shown alongside corresponding $\alpha$-$\beta$ and XNOR-Net filters. Orange signifies the highest response areas. We can see that DAB-Net has significantly better responses when compared to XNOR-Net}
        \label{fig:introdiagram}
        \vspace{-0.4cm}
\end{figure}
 
Later works have tended to move away from binary representations of weights/inputs to multi-bit representations. The reason for this was mainly the large accuracy degradation observed in binary networks. While some works \cite{tang2017train} have proposed methods to recover some of the lost accuracy, this leads to the natural question of whether, in theory, binary-representations of neural networks can be used at all to effectively approximate a full-precision network. If shown to be sufficient, the search for an optimally accurate binarization technique is worthwhile, due to the large gains in speedups (due to binary operations rather than full-prec MACs) and compression compared to multi-bit representations. 

In our paper, we make the following contributions:
\begin{enumerate}\vspace{-0.2cm}
\item We show that binary representations are as expressive as full precision neural networks for polynomial functions, and offer theoretical insights into the same. \vspace{-0.25cm}
\item We present a generalized, distribution-aware representation for binary networks, and proceed to calculate the generalized parameter-values for any binary network. \vspace{-0.25cm}
\item We offer an intuitive analysis and comparison of our representation vis-a-vis previous representations, as  illustrated in Figure \ref{fig:weightdistributionsample}. \vspace{-0.25cm}
\item We provide a provably efficient implementation of networks trained using this representation. \vspace{-0.25cm}
\item We demonstrate the effectiveness of our method by extensive experiments applying it to popular model architectures on large-scale sketch datasets and improving upon existing binarization approaches. \vspace{-0.1cm}
\end{enumerate}

We also offer intuitions about how this technique might be effective in problems involving data that is inherently binary, such as sketches, as shown in Figure \ref{fig:introdiagram}. Sketches are a universal form of communication and are easy to draw through mobile devices - thus emerging as a new paradigm with interesting areas to explore, such as fast classification and sketch-based image retrieval.

{\bf Reproducibility:} Our implementation can be found on GitHub \footnote{https://github.com/erilyth/DistributionAwareBinarizedNetworks-WACV18}

\section{Related Work}
We ask the question: Do CNNs need the representational power of 32-bit floating point operations, especially for binary-valued data such as sketches? Is it possible to cut down memory costs and make output computations significantly less expensive? In recent years, several different approaches were proposed to achieve network compression and speedups, and special-purpose networks were proposed for sketch classification/retrieval tasks. These are summarized below:

{\bf Sketch Recognition:} 
Many deep-network based works in the past did not lead to fruitful results before, primarily due to these networks being better suited for images rather than sketches.
Sketches have significantly different characteristics as compared to images, and require specialized, fine-tuned networks to work with. Sketch-a-Net from Yu \etal. \cite{yu2015sketch} took these factors into account, and proposed a carefully designed network structure that suited sketch representations. Their single-model showed tremendous increments over the then state-of-the-art, and managed to beat the average human performance using a Bayesian Fusion ensemble. Being a significant achievement in this problem - since beating human accuracy in recognition problems is difficult - this model has been adopted by a number of later works {Bui \etal. \cite{bui2016sbir}, Yu \etal \cite{yu2016shoe}, Wang \etal \cite{wang2016crossir}}.
\begin{table}[t]
\centering
\begin{tabular}{|l|c|}
\hline
{\bf Method} &  {\bf Compression} \\
\hline
Finetuned SVD 2 \cite{yang2015deep} & 2.6x \\
Circulant CNN 2 \cite{cheng2015exploration} & 3.6x \\
Adaptive Fastfood-16 \cite{yang2015deep} & 3.7x \\
Collins \etal \cite{collins2014memory} & 4x \\
Zhou \etal \cite{zhou2016less} & 4.3x \\
ACDC \cite{moczulski2015acdc} & 6.3x \\
Network Pruning \cite{han2015deep} & 9.1x \\
Deep Compression \cite{han2015deep} & 9.1x \\
GreBdec \cite{yu2017compressing} & 10.2x \\
Srinivas \etal \cite{srinivas2017training} & 10.3x \\
Guo \etal \cite{guo2016dynamic} & 17.9x \\
\hline
{\bf Binarization} & {\bf ${\approx}$32x} \\ 
\hline
\end{tabular}
\caption{Comparison of Binarization and other methods in terms of compression.}
\label{table:versions_typesofcompression}
\end{table}

{\bf Pruning Networks for Compression}: 
Optimal Brain Damage \cite{Cunn1} and Optimal Brain Surgeon \cite{Hassibi} introduced a network pruning technique based on the Hessian of the loss function. Deep Compression \cite{han2015deep} also used pruning to achieve compression by an order of magnitude in various standard neural networks. It further reduced non-runtime memory by employing trained quantization and Huffman coding. Network Slimming \cite{liu2017learning} introduced a new learning scheme for CNNs that leverages channel-level sparsity in networks, and showed compression and speedup without accuracy degradation, with decreased run-time memory footprint as well. 
We train our binary models from scratch, as opposed to using pre-trained networks as in the above approaches.

{\bf Higher Bit Quantization:} 
HashedNets \cite{chen2015hashed} hashed network weights to bin them. Zhou et al. \cite{zhou2017inq} quantized networks to 4-bit weights, achieving 8x memory compression by using 4 bits to represent 16 different values and 1 bit to represent zeros. Trained Ternary Quantization \cite{zhu2016ternary} uses 2-bit weights and scaling factors to bring down model size to 16x compression, with little accuracy degradation. Quantized Neural Networks\cite{hubara2016quantized} use low-precision quantized weights and inputs and replaces arithmetic operations with bit-wise ones, reducing power consumption. DoReFa-Net \cite{zhou2016dorefa} used low bit-width gradients during backpropagation, and obtained train-time speedups. Ternary Weight Networks \cite{li2016ternary} optimize a threshold-based ternary function for approximation, with stronger expressive abilities than binary networks.  
The above works cannot leverage the speedups gained by XNOR/Pop-count operations which could be performed on dedicated hardware, unlike in our work. This is our primary motivation for attempting to improve binary algorithms.

{\bf Binarization:} 
We provide an optimal method for calculating binary weights, and we show that all of the above binarization techniques were special cases of our method, with less accurate approximations. Previous binarization papers performed binarization independent of the distribution weights, for example \cite{rastegari2016xnor}. The method we introduce is distribution-aware, i.e. looks at the distribution of weights to calculate an optimal binarization.  

BinaryConnect \cite{courbariaux2015binaryconnect} was one of the first works to use binary (+1, -1) values for network parameters, achieving significant compression. XNOR-Nets \cite{rastegari2016xnor} followed the work of BNNs \cite{hubara2016binarized}, binarizing both layer weights and inputs and multiplying them with scaling constants - bringing significant speedups by using faster XNOR-Popcount operations to calculate convolutional outputs. Recent research proposed a variety of additional methods - including novel activation functions \cite{cai2017deep}, alternative layers \cite{tang2017train}, approximation algorithms \cite{hou2016loss}, fixed point bit-width allocations \cite{pmlr-v48-linb16}. Merolla et al. \cite{DBLP:journals/corr/MerollaAAEM16} and Anderson et al. \cite{anderson2017high} offer a few theoretical insights and analysis into binary networks. Further works have extended this in various directions, including using local binary patterns \cite{juefei2016local} and lookup-based compression methods \cite{bagherinezhad2016lcnn}.

\section{Representational Power of Binary Networks} 
Many recent works in network compression involve higher bit weight quantization using two or more bits \cite{zhou2017inq,zhu2016ternary,li2016ternary} instead of binarization, arguing that binary representations would not be able to approximate full-precision networks. In light of this, we explore whether the representational power that binary networks can offer is theoretically sufficient to get similar representational power as full-precision networks.

Rolnick \etal \cite{lin2017does,rolnick2017power} have done extensive work in characterizing the expressiveness of neural networks. They claim that due to the nature of functions - that they depend on real-world physics, in addition to mathematics - the seemingly huge set of possible functions could be approximated by deep learning models. From the Universal Approximation Theorem \cite{cybenko1989approximation}, it is seen that any arbitrary function can be well-approximated by an Artificial Neural Network; but \textit{cheap learning}, or models with far fewer parameters than generic ones, are often sufficient to approximate multivariate monomials - which are a class of functions with practical interest, occurring in most real-world problems.

We can define a binary neural network having $k$ layers with activation function $\sigma(x)$  and consider how many neurons are required to compute a multivariate monomial $p(x)$ of degree $d$. The network takes an $n$ dimensional input $\mathbf{x}$, producing a one dimensional output $p(x)$. We define $B_k(p,\sigma)$ to be the minimum number of binary neurons (excluding input and output) required to approximate $p$, where the error of approximation is of degree at least $d+1$ in the input variables. For instance, $B_1(p,\sigma)$ is the minimal integer $m$ such that:
$$\sum_{j=1}^m w_j \sigma\left(\sum_{i=1}^n a_{ij} x_i\right) = p(x) + \mathcal{O}(x_1^{d + 1} + \ldots + x_n^{d + 1}).$$ 
Any polynomial can be approximated to high precision as long as input variables are small enough \cite{lin2017does}. Let $B(p, \sigma) = \min_{k \geq 0} B_k(p, \sigma)$.

\begin{theorem}
\label{theorem:binaryrepresentation}
For $p(\mathbf{x})$ equal to the product $x_1x_2\cdots x_n$, and for any $\sigma$ with all nonzero Taylor coefficients, we have one construction of a binary neural network which meets the condition
\begin{equation}
B_k(p, \sigma) = \mathcal{O}\left(n^{(k-1)/k}\cdot 2^{n^{1/k}}\right).\label{eqn:constantlayers}
\end{equation}

Proof of the above can be found in the supplementary material. 
\end{theorem}

Conjecture III.2. of Rolnick \etal \cite{rolnick2017power} says that this bound is approximately optimal. If this conjecture proves to be true, weight-binarized networks would have the same representational power as full-precision networks, since the network that was essentially used to prove that the above theorem - that a network exists that can satisfy that bound - was a binary network. 

The above theorem shows that any neural network that can be represented as a multivariate polynomial function is considered as a simplified model with ELU-like activations, using continuously differentiable layers - so pooling layers are excluded as well. While there can exist a deep binary-weight network that can possibly approximate polynomials similar to full precision networks, it does say that such a representation would be efficiently obtainable through Stochastic Gradient Descent. Also, this theorem assumes only weights are binarized, not the activations. Activation binarization typically loses a lot of information and might not be a good thing to do frequently. However, this insight motivates the fact that more investigation is needed into approximating networks through binary network structures.

\section{Distribution-Aware Binarization}

We have so far established that binary representations are possibly sufficient to approximate a polynomial with similar numbers of neurons as a full-precision neural network. We now investigate the question - What is the most general form of binary representation possible? In this section, we derive a generalized distribution-aware formulation of binary weights, and provide an efficient implementation of the same. We consider models binarized with our approach as DAB-Nets (Distribution Aware Binarized Networks).

We model the loss function layer-wise for the network. We assume that inputs to the convolutional layers are binary - i.e. belong to $\{+1, -1\}$, and find constants $\alpha$ and $\beta$ (elaborated below) as a general binary form for layer weights. These constants are calculated from the distribution of real-valued weights in a layer - thus making our approach \textit{distribution-aware}. 

\subsection{Derivation}
Without loss of generality, we assume that $\mathbf{W}$ is a vector in $R^{n}$ , where $n = c\cdot w\cdot h$.
We attempt to binarize the weight vector $\mathbf{W}$ to $\widetilde{\mathbf{W}}$ which takes a form similar to this example - $[\alpha \alpha... \beta \alpha \beta]$. Simply put, $\widetilde{\mathbf{W}}$ is a vector consisting of scalars $\alpha$ and $\beta$, the two values forming the binary vector. We represent this as $\widetilde{\mathbf{W}} = \alpha \mathbf{e} + \beta \mathbf{(1-e)}$ where $\mathbf{e}$ is a vector such that $\mathbf{e} \in \{0,1\}^n \ni \mathbf{e} \neq 0$ and $\mathbf{e} \neq 1$.  We define $K$ as $\mathbf{e}^T\mathbf{e}$ which represents the number of ones in the $\mathbf{e}$ vector. Our objective is to find the best possible binary approximation for $\mathbf{W}$. We set up the optimization problem as: $$\widetilde{\mathbf{W}}^\ast = \underset{\widetilde{\mathbf{W}}}{\mathrm{argmin}}\mid\mid \mathbf{W}-\widetilde{\mathbf{W}}\mid\mid^{2}$$ We formally state this as the following: \\

\label{approx}
{\it The optimal binary weight vector $\widetilde{\mathbf{W}}^\ast$ for any weight vector $\mathbf{\mathbf{W}}$ which minimizes the approximate-error function $\mathbf{J} = \mid\mid \mathbf{W}-\widetilde{\mathbf{W}}\mid\mid^{2}$ can be represented as: 
$$\widetilde{\mathbf{W}}^\ast = \alpha \mathbf{e} + \beta \mathbf{(1-e)}  \; where\\$$
$$ \alpha =\frac{\mathbf{W}^{T}\mathbf{e}}{K} \;, \; \beta = \frac{\mathbf{W}^{T}\mathbf{(1-e)}}{n-K} $$ for a given $K$. That is, given a $K$, the optimal selection of $\mathbf{e}$ would correspond to either the $K$ smallest weights of $\mathbf{W}$ or the $K$ largest weights of $\mathbf{W}$. 

The best suited $K$, we calculate the value of the following expression for every value of $K$, giving us an $\mathbf{e}$, and maximize the expression: $$  \mathbf{e}^\ast  = \underset{\mathbf{e}}{\mathrm{argmax}}  (\frac{\mid\mid\mathbf{\mathbf{W}}^T\mathbf{e}\mid\mid^2}{K}+\frac{\mid\mid\mathbf{W}^T\mathbf{(1-e)}\mid\mid^2}{n-K}) \\$$
A detailed proof of the above can be found in the supplementary material.}
\\\\
The above representation shows the values obtained for $\mathbf{e}$, $\alpha$ and $\beta$ are the optimal approximate representations of the weight vector $\mathbf{W}$. The vector $\mathbf{e}$, which controls the number and distribution of occurrences of $\alpha$ and $\beta$, acts as a mask of the top/bottom $K$ values of $\mathbf{W}$. We assign $\alpha$ to capture the greater of the two values in magnitude. Note that the scaling values derived in the XNOR formulation, $\alpha$ and $-\alpha$, are a special case of the above, and hence our approximation error is at most that of the  XNOR error. We explore what  this function represents and how this relates to previous binarization techniques in the next subsection.

\subsection{Intuitions about DAB-Net}

In this section, we investigate intuitions about the derived representation. We can visualize that $\mathbf{e}$ and $\mathbf{(1-e)}$ are orthogonal vectors. Hence, if normalized, $\mathbf{e}$ and $\mathbf{(1-e)}$ form a basis for a subspace $R^{2}$. Theorem 2 says the best $\alpha$ and $\beta$ can be found by essentially projecting the weight matrix $\mathbf{W}$ into this subspace, finding the vector in the subspace which is \textit{closest} to $\mathbf{e}$ and $\mathbf{(1-e)}$ respectively. $$\alpha = \frac{\langle \mathbf{W}, \mathbf{e} \rangle}{\langle \mathbf{e} , \mathbf{e} \rangle} \cdot \mathbf{e} \ , \  \beta = \frac{\langle \mathbf{W}, \mathbf{(1-e)} \rangle}{\langle \mathbf{(1-e)} , \mathbf{(1-e)} \rangle} \cdot \mathbf{(1-e)}$$  

We also show that our derived representation is different from the previous binary representations since we cannot derive them by assuming a special case of our formulation. XNOR-Net \cite{rastegari2016xnor} or BNN \cite{hubara2016binarized}-like representations cannot be obtained from our formulation. However, in practice, we are able to simulate XNOR-Net by constraining $\mathbf{W}$ to be mean-centered and $K = \frac{n}{2}$, since roughly half the weights are above 0, the other half below, as seen in Figure \ref{fig:kacrosslayers} in Section \ref{sec:kacrosslayers}.

\subsection{Implementation}

\begin{algorithm}[t]
\caption{ Finding an optimal K value. }
\begin{algorithmic}[1]
\State \texttt{Initialization}
\State $\mathbf{W}$ = 1D weight vector
\State $T$ = Sum of all the elements of $\mathbf{W}$
\State  Sort($\mathbf{W}$)
\State $D$ = $[0 0... 0]$ ~~//~{Empty array of same size as $\mathbf{W}$}
\State $optK_{1}$ = 0 ~~//~{Optimal value for K}
\State $maxD_{1}$ = 0 ~~//~{Value of D for optimal K value} \\

\For{$I$= 1 to D.size}
	\State $P_{i} = P_{i-1} + \mathbf{W}_{i}$
	\State $D_{i} = \frac{P_{i}^{2}}{i} + \frac{(T-P_{i})^{2}}{n-i}$
    \If{$D_{i} \geq maxD_{1}$}
    \State $maxD_{1}$ = $D_{i}$
    \State $optK_{1}$ = i
    \EndIf
\EndFor
\\
\State Sort($\mathbf{W}$, reverse=true) and {\bf Repeat} steps 4-13 with $optK_{2}$ and $maxD_{2}$
\\
\State $optK_{final}$ = $optK_{1}$
\If{$maxD_{2} > maxD_{1}$}
\State $optK_{final}$ = $optK_{2}$
\EndIf
\\
\State {\bf return} $optK_{final}$
\end{algorithmic}
\label{alg:partitionalgo}
\end{algorithm}

The representation that we earlier derived requires to be efficiently computable, in order to ensure that our algorithm runs fast enough to be able to train binary networks. In this section, we investigate the implementation, by breaking it into two parts: 1) Computing the parameter $K$ efficiently for every iteration. 2) Training the entire network using that value of $K$ for a given iteration. We show that it is possible to get an efficiently trainable network at minimal extra cost. We provide an efficient algorithm using Dynamic Programming which computes the optimal value for $K$ quickly at every iteration. \\

\subsubsection{Parallel Prefix-Sums to Obtain $K$}
\begin{theorem}
\label{theorem:approx}
The optimal $K^*$ which minimizes the value $\mathbf{e}$ can be computed in $O(n \cdot logn)$ complexity.
\end{theorem}

Considering one weight filter at a time for each convolution layer, we flatten the weights into a 1-dimensional weight vector $\mathbf{W}$. We then sort the vector in ascending order and then compute the prefix-sum array $P$ of $\mathbf{W}$. For a selected value of $K$, the term to be maximized would be $(\frac{\mid\mid\mathbf{\mathbf{W}}^T\mathbf{e}\mid\mid^2}{K}+\frac{\mid\mid\mathbf{W}^T\mathbf{(1-e)}\mid\mid^2}{n-K})$, which is equal to $(\frac{P_{i}^{2}}{i} + \frac{(T-P_{i})^{2}}{n-i})$ since the top $K$ values in $\mathbf{W}$ sum up to $P_{i}$ where $T$ is the sum of all weights in $\mathbf{W}$. We also perform the same computation with a descending order of $\mathbf{W}$'s weights since $K$ can correspond to either the smallest $K$ weights or the largest $K$ weights as we mentioned earlier. In order to speed this up, we perform these operations on all the weight filters at the same time considering them as a 2D weight vector instead. Our algorithm runs in $O(n \cdot logn)$ time complexity, and is specified in Algorithm \ref{alg:partitionalgo}. This algorithm is integrated into our code, and will be provided alongside.

\subsubsection{Forward and Backward Pass}
Now that we know how to calculate $K$, $\mathbf{e}$, $\alpha$, and $\beta$ for each filter in each layer optimally, we can compute $\widetilde{\mathbf{W}}$ which approximates $\mathbf{W}$ well. Here, $topk(\mathbf{W},K)$ represents the top $K$ values of $\mathbf{W}$ which remain as is whereas the rest are converted to zeros. Let $\mathbf{T_k} = topk(\mathbf{W}, K)$.

\begin{corollary}[Weight Binarization]\label{corollary:forward}
The optimal binary weight $\widetilde{\mathbf{W}}$ can be represented as,
\[
\widetilde{\mathbf{W}} = \alpha.sgn(\mathbf{T_k}) + \beta.(1-sgn(\mathbf{T_k}))
\]
where,
\[
\alpha = \frac{\mathbf{T_k}}{K} \ and \ 
\beta = \frac{(\mathbf{W}-\mathbf{T_k})}{n-K}
\]
\end{corollary}

Once we have $\widetilde{\mathbf{W}}$, we can perform convolution as $\bf{I} \circledast \widetilde{\mathbf{W}}$ during the forward pass of the network. Similarly, the optimal gradient $\widetilde{\mathbf{G}}$ can be computed as follows, which is back-propagated throughout the network in order to update the weights:

\begin{algorithm}[t]
{
  \caption{Training an $L$-layers CNN with binary weights:}
  \label{alg:trainbinconv}       
  \begin{algorithmic}[1]
  \State A minibatch of inputs and targets  ($\bI, \bY$), cost function $C(\bY,\hat{\bY})$, current weight $\mathbf{W}^t$ and current learning rate $\eta^t$.     
  \State updated weight $\mathbf{W}^{t+1}$ and updated learning rate $\eta^{t+1}$. 
  \State {\bf Binarizing weight filters}:
  \State $\mathbf{W}^t$ = MeanCenter($\mathbf{W}^t$)
  \State $\mathbf{W}^t$ = Clamp($\mathbf{W}^t$, -1, 1)
  \State $\mathbf{W}_{real}$ = $\mathbf{W}^t$
  \For{$l=1$ to $L$}
      \For{$j^{\text{th}}$ filter in $l^{\text{th}}$ layer}
      	  \State Find $K_{lj}$ using Algorithm \ref{alg:partitionalgo}
          \State $\alpha_{lj}=\frac{topk(\mathbf{W}_{lj},K_{lj})}{K_{lj}}$
          \State $\beta_{lj}=-\frac{(\mathbf{W}_{lj}-topk(\mathbf{W}_{lj},K_{lj}))}{n-K_{lj}}$
          \State $\widetilde{\mathbf{W}}_{lj}=\alpha.sgn(topk(\mathbf{W}_{lj},K_{lj}))$ \\ \ \ \ \ \ \ \ \ \ \ \ \ \ \ \ \ \ \ \ \ \ \ \  $ + \ \beta.(1-sgn(topk(\mathbf{W}_{lj},K_{lj})))$
      \EndFor
   \EndFor
   \\
   \State $\hat{\bY}=$ ~~\textbf{BinaryForward}$(\bI,\widetilde{\mathbf{W}})$
   \\
   \State $\frac{\partial C}{\partial \widetilde{\mathbf{W}}} =$ \textbf{BinaryBackward}$(\frac{\partial C}{\hat{\bY}}, \widetilde{\mathbf{W}})$ ~~//~{ Standard backward propagation except that gradients are computed using $\widetilde{\mathbf{W}}$ instead of $\mathbf{W}^t$} as mentioned in Theorem. \ref{theorem:backward}
   \\
 \State We then copy back the real weights in order to apply the gradients computed. $\mathbf{W}^t$ = $\mathbf{W}_{real}$ \\
 \State $\mathbf{W}^{t+1}$ = \textbf{UpdateParameters}$(\mathbf{W}^{t},\frac{\partial C}{\partial \widetilde{\mathbf{W}}}, \eta^t)$ 
 \State $\eta^{t+1}=$ \textbf{UpdateLearningrate}$(\eta^t, t)$
  \end{algorithmic}
  }
\end{algorithm}

\begin{theorem}[Backward Pass]\label{theorem:backward}
The optimal gradient value $\widetilde{G}$ can be represented as,
\begin{dmath}
\widetilde{\mathbf{G}} = \widetilde{\mathbf{G_1}} + \widetilde{\mathbf{G_2}}
\end{dmath}
where,
\begin{dmath}
\widetilde{\mathbf{G_1}} = \frac{sgn(\mathbf{T_k})}{K}\circ sgn(\mathbf{T_k}) + \frac{||\mathbf{T_k}||_{l1}}{K} .STE(\mathbf{T_k})
\end{dmath}
\begin{dmath}
\widetilde{\mathbf{G_2}} = \frac{sgn(\mathbf{W}-\mathbf{T_k})}{n-K} \circ (1-sgn(\mathbf{T_k})) + \frac{||\mathbf{W}-\mathbf{T_k}||_{l1}}{n-K}.STE(\mathbf{W} - \mathbf{T_k})
\end{dmath}
\begin{equation}
STE(\mathbf{T_k})^{i} = 
    \begin{cases}
      \mathbf{T_k}^{i}, \text{where} \ |\mathbf{W}|^{i}<=1 \\
      0,\ \text{elsewhere}
    \end{cases}
\end{equation}
\end{theorem}

The gradient vector, as seen above, can be intuitively understood if seen as the sum of two independent gradients $\widetilde{\mathbf{G_1}}$ and $\widetilde{\mathbf{G_2}}$, each corresponding to the vectors $\mathbf{e}$ and $\mathbf{(1-e)}$ respectively. Further details regarding the derivation of this gradient would be provided in the supplementary material.


\subsection{Training Procedure}

Putting all the components mentioned above together, we have outlined our training procedure in Algorithm \ref{alg:trainbinconv}. During the forward pass of the network, we first mean center and clamp the current weights of the network. We then store a copy of these weights as $\mathbf{W}_{real}$. We compute the binary forward pass of the network, and then apply the backward pass using the weights $\widetilde{\mathbf{W}}$, computing gradients for each of the weights. We then apply these gradients on the original set of weights $\mathbf{W}^t$ in order to obtain $\mathbf{W}^{t+1}$. In essence, binarized weights are used to compute the gradients, but they are applied to the original stored weights to perform the update. This requires us to store the full precision weights during training, but once the network is trained, we store only the binarized weights for inference.

\section{Experiments}
We empirically demonstrate the effectiveness of our optimal distribution-aware binarization algorithm (DAB-Net) on the TU-Berlin and Sketchy datasets. We compare DAB-Net with BNN and XNOR-Net \cite{rastegari2016xnor} on various architectures, on two popular large-scale sketch recognition datasets as sketches are sparse and binary. Also, they are easier to train with than standard images, for which we believe the algorithm needs to be stabilized - in essence, the $K$ value must be restricted to change by only slight amounts. We show that our approach is superior to existing binarization algorithms, and can generalize to different kinds of CNN architectures on sketches.

\subsection{Experimental Setup}
In our experiments, we define the network having only the convolutional layer weights binarized as WBin, the network having both inputs and weights binarized as FBin and the original full-precision network as FPrec. Binary Networks have achieved accuracies comparable to  full-precision networks on limited domain/simplified  datasets like CIFAR-10, MNIST, SVHN, but show considerable losses on larger datasets. Binary networks are well suited for sketch data due to its binary and sparse nature of the data. 

{\bf TU-Berlin:} The TU-Berlin \cite{eitz2012hdhso} dataset is the most popular large-scale free-hand sketch dataset containing sketches of 250 categories, with a human sketch-recognition accuracy of 73.1\% on an average.

{\bf Sketchy:} A recent large-scale free-hand sketch dataset containing 75,471 hand-drawn sketches spanning 125 categories. This dataset was primarily used to cross-validate results obtained on the TU-Berlin dataset, to ensure the robustness of our approach with respect to the method of data collection.

For all the datasets, we first resized the input images to 256 x 256. A 224 x 224 (225 x 225 for Sketch-A-Net) sized crop was then randomly taken from an image with standard augmentations such as rotation and horizontal flipping, for TU-Berlin and Sketchy. In the TU-Berlin dataset, we use three-fold cross validation which gives us a 2:1 train-test split ensuring that our results are comparable with all previous methods. For Sketchy, we use the training images for retrieval as the training images for classification, and validation images for retrieval as the validation images for classification. We report ten-crop accuracies on both the datasets.

We used the PyTorch framework to train our networks. We used the Sketch-A-Net\cite{yu2015sketch}, ResNet-18\cite{he2016deep} and GoogleNet\cite{43022} architectures. Weights of all layers except the first were binarized throughout our experiments, except in Sketch-A-Net for which all layers except first and last layers were binarized.  All networks were trained from scratch. We used the Adam optimizer for all experiments. Note that we do not use a bias term or weight decay for binarized Conv layers. We used a batch size of 256 for all Sketch-A-Net models and a batch size of 128 for ResNet-18 and GoogleNet models, the maximum size that fits in a 1080Ti GPU. Additional experimental details are available in the supplementary material.

\begin{table}[t]  
\resizebox{\columnwidth}{!}{
\begin{tabular}{|l|c|c|c|}
\hline
\multirow{2}{*}{\bf Models} &  \multirow{2}{*}{\bf Method} &  \multicolumn{2}{c|}{\sc { \bf Accuracies}}\\
\cline{3-4}

 &   & TU-Berlin & Sketchy\\
\hline
\multirow{5}{*}{Sketch-A-Net} & FPrec  & 72.9\%  & 85.9\%\\
 & WBin (BWN)  & 73.0\% & 85.6\%\\
 & FBin (XNOR-Net) & 59.6\% & 68.6\% \\
 & WBin DAB-Net  & 72.4\% & 84.0\% \\
 & FBin DAB-Net  & {\bf 60.4\%} & {\bf 70.6\%} \\
\hline
Improvement & XNOR-Net vs DAB-Net & \textcolor{darkspringgreen}{+0.8\%} & \textcolor{darkspringgreen}{+2.0\%}\\
\hline
\multirow{5}{*}{ResNet-18} & FPrec & 74.1\% & 88.7\% \\
 & WBin (BWN) & 73.4\%  & 89.3\%\\
 & FBin (XNOR-Net) & 68.8\% & 82.8\%\\
 & WBin DAB-Net  & 73.5\% & 88.8\%\\
 & FBin DAB-Net  & {\bf 71.3\%} & {\bf 84.2\%}\\
\hline
Improvement & XNOR-Net vs DAB-Net & \textcolor{darkspringgreen}{+2.5\%} & \textcolor{darkspringgreen}{+1.4\%}\\
\hline
\multirow{5}{*}{GoogleNet} & FPrec & 75.0\% & 90.0\% \\
 & WBin (BWN) & 74.8\%  & 89.8\%\\
 & FBin (XNOR-Net) & 72.2\% & 86.8\% \\
 & WBin DAB-Net  & 75.7\% & 90.1\%\\
 & FBin DAB-Net  & {\bf 73.7\%} & {\bf 87.4\%} \\
\hline
Improvement & XNOR-Net vs DAB-Net & \textcolor{darkspringgreen}{+1.5\%} & \textcolor{darkspringgreen}{+0.6\%}\\
\hline
\end{tabular}}
\caption{Our DAB-Net models compared to FBin, WBin and FPrec models on TU-Berlin and Sketchy in terms of accuracy.} 
\label{table:tub_recacc}
\vspace{-0.5cm}
\end{table}

\begin{figure*}[t]
\vspace*{-0.5cm}
\begin{center}
\begin{tabular}{cc}
           \includegraphics[page=1,width=0.9\columnwidth]{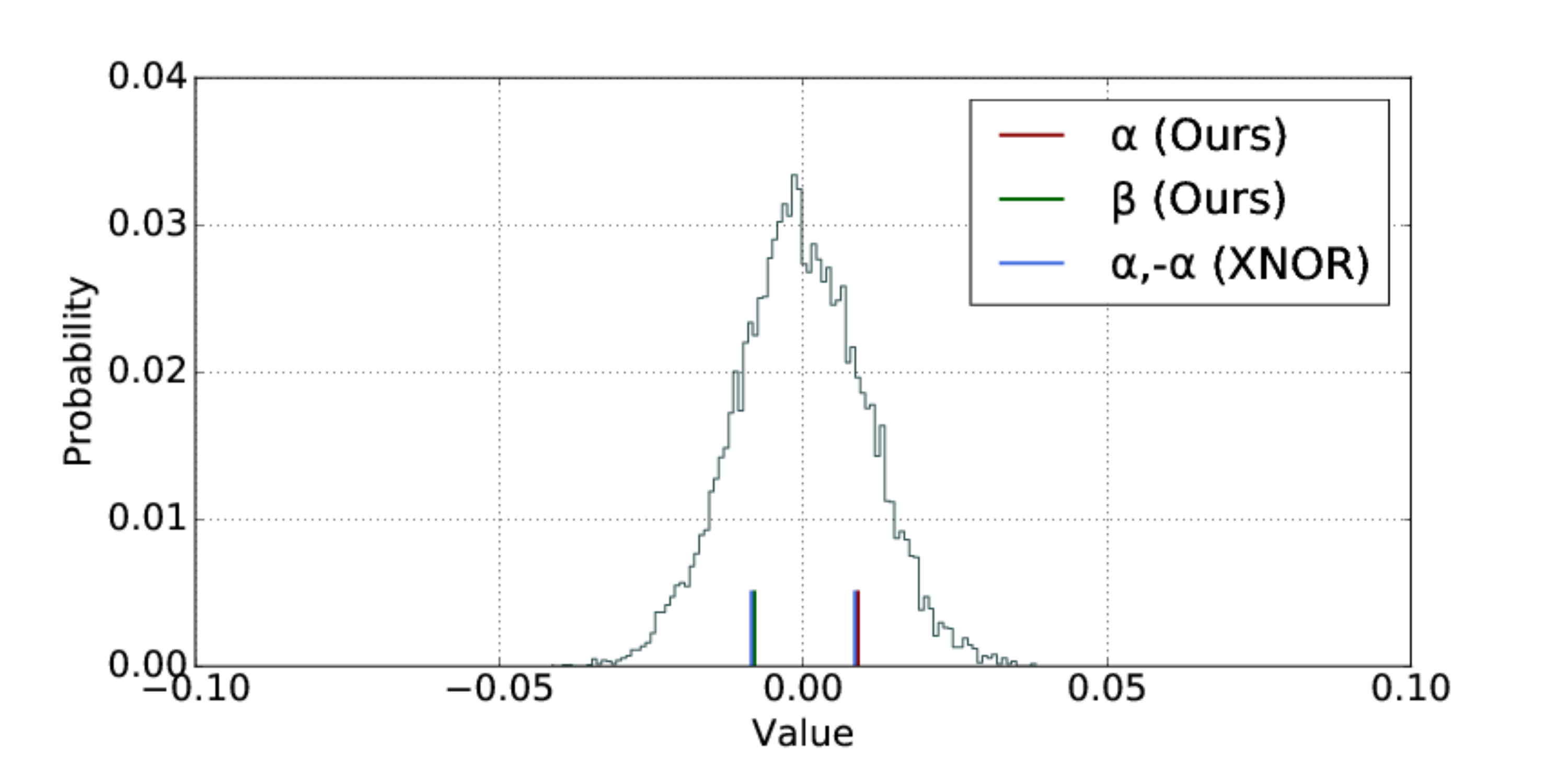} & 
           \includegraphics[page=2,width=0.9\columnwidth]{./figure21222324.pdf}\\
           (1) & (2)\\
           \includegraphics[page=3,width=0.9\columnwidth]{./figure21222324.pdf} & \includegraphics[page=4,width=0.9\columnwidth]{./figure21222324.pdf}\\
           (3) & (4)\\  
\end{tabular}
\end{center}
\vspace*{-0.5cm}
\caption{Sub-figures (1) to (4) show the train-time variation of $\alpha$ and $\beta$ for a layer filter. Initially, $\alpha$ and $\beta$ have nearly equal magnitudes, similar to the XNOR-Net formulation, but as we progress to (4), we see that $\alpha$ and $\beta$ have widely different magnitudes.
Having just one scaling constant (XNOR-Net) would be a comparatively poor approximator.}
        \label{fig:alphabetaovertime}
\end{figure*}

\subsection{Results}

We compare the accuracies of our distribution aware binarization algorithm for WBin and FBin models on the TU-Berlin and Sketchy datasets. Note that higher accuracies are an improvement, hence stated in green in Table \ref{table:tub_recacc}.
On the TU-Berlin and Sketchy datasets in Table \ref{table:tub_recacc}, we observe that FBin DAB-Net models consistently perform better over their XNOR-Net counterparts. They improve upon XNOR-Net accuracies by 0.8\%, 2.5\%, and 1.5\% in Sketch-A-Net, ResNet-18, and GoogleNet respectively on the TU-Berlin dataset. Similarly, they improve by 2.0\%, 1.4\%, and 0.6\% respectively on the Sketchy dataset. We also compare them with state-of-the-art sketch classification models in Table \ref{table:sketchcomp}. We find that our compressed models perform significantly better than the original sketch models and offer compression, runtime and energy savings additionally.
\begin{figure}[t]
\vspace{-0.4cm}
    \begin{minipage}[b]{0.9\linewidth}
    \centering
           \includegraphics[width=1\textwidth]{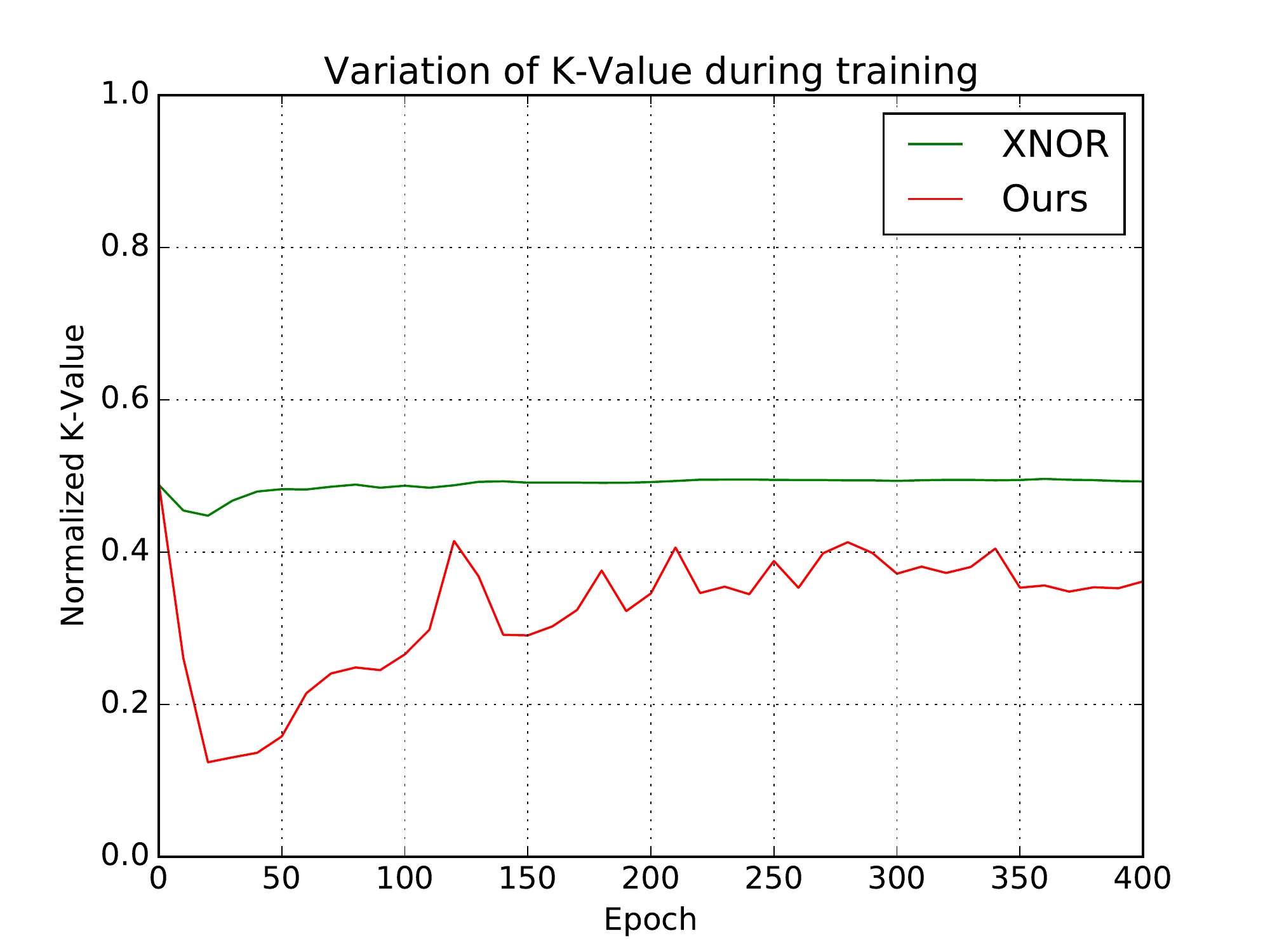}
           \caption{The variation of the normalized K-value over time during training. It falls initially but converges eventually to 0.35. The normalized K-value for XNOR-Net remains almost at 0.5 till the end.}
        \label{fig:kvalovertime}
    \end{minipage}
    \vspace{-0.4cm}
\end{figure}
\begin{table}[t]
\begin{center}
\begin{tabular}{|l|c|c|c|l|}
\hline
{\bf Models}  &  {\bf Accuracy}\\
\hline
AlexNet-SVM  & 67.1\%\\
AlexNet-Sketch  & 68.6\%\\
Sketch-A-Net SC  & 72.2\%\\
Humans & {73.1\%}\\
Sketch-A-Net-2\footnotemark \cite{yu2017sketch} & {\bf 77.0\%}\\
\hline
Sketch-A-Net WBin DAB-Net & 72.4\%\\
ResNet-18 WBin DAB-Net & 73.5\%\\
GoogleNet WBin DAB-Net & {\bf 75.7\%}\\
\hline
Sketch-A-Net FBin DAB-Net & 60.4\%\\
ResNet-18 FBin DAB-Net & 71.3\%\\
GoogleNet FBin DAB-Net & {\bf 73.7\%}\\
\hline
\end{tabular}
\end{center}
\vspace{-0.4cm}
\caption{A comparison between state-of-the-art single model accuracies of recognition systems on the TU-Berlin dataset.}
\vspace{-0.4cm}
\label{table:sketchcomp}
\end{table}
\footnotetext{It is the sketch-a-net SC model trained with additional imagenet data, additional data augmentation strategies and considering an ensemble, hence would not be a direct comparison}

Our DAB-Net WBin models attain accuracies similar to BWN WBin models and do not offer major  improvements mainly because WBin models achieve FPrec accuracies already, hence do not have much scope for improvement unlike FBin models. Thus, we conclude that our DAB-Net FBin models are able to attain significant accuracy improvements over their XNOR-Net counterparts when everything apart from the binarization method is kept constant.

\subsection{XNOR-Net vs DAB-Net}

We measure how $K$, $\alpha$, and $\beta$ vary across various layers over time during training, and these variations are observed to be quite different from their corresponding values in XNOR-Net. These observations show that binarization can approximate a network much better when it is distribution-aware (like in our technique) versus when it is distribution-agnostic (like XNOR-Nets).

\subsubsection{Variation of $\alpha$ and $\beta$ across Time}

We plot the distribution of weights of a randomly selected filter belonging to a layer and observe that $\alpha$ and $\beta$ of DAB-Net start out to be similar to $\alpha$ and $-\alpha$ of XNOR-Nets, since the distributions are randomly initialized. However, as training progresses, we observe as we go from Subfigure (1) to (4) in Figure \ref{fig:alphabetaovertime}, the distribution eventually becomes non-symmetric and complex, hence our values  significantly diverge from their XNOR-Net counterparts. This divergence signifies a better approximation of the underlying distribution of weights in our method, giving additional evidence to our claim that the proposed DAB-Net technique gives a better representation of layer weights, significantly different from that of XNOR-Nets. 

\subsubsection{Variation of $K$ across Time and Layers}\label{sec:kacrosslayers}

We define \textit{normalized} $K$ as the $\frac{K}{n}$ for a layer filter. For XNOR-Nets, $K$ would be the number of values below zero in a given weight filter - which has minimal variation, and does not take into consideration the distribution of weights in the filter - as $K$ in this case is simply the number of weights below a certain fixed global threshold, zero. However, we observe that the $K$ computed in DAB-Net varies significantly across epochs initially, but slowly converges to an optimal value for the specific layer as shown in Figure \ref{fig:kvalovertime}.

\begin{figure}[t]
\vspace{-0.4cm}
\centering
		\includegraphics[width=0.9\columnwidth]{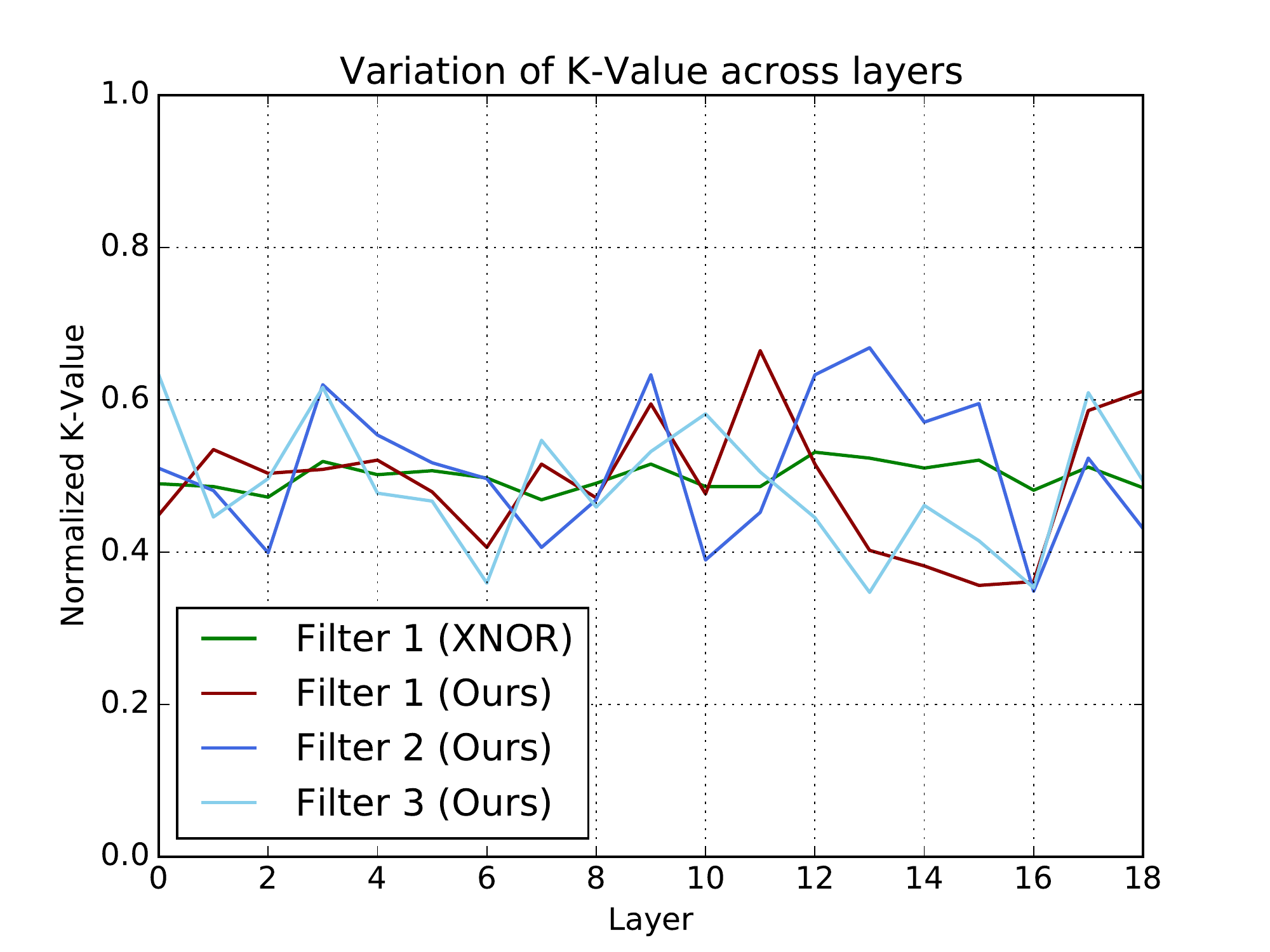}
        \caption{The variation of normalized K values on random filters across layers. The K-value corresponding to DAB-Net varies across layers based on the distribution of weights of the specific layer, which is not captured by XNOR-Net.}
        \label{fig:kacrosslayers}
\end{figure}

We also plot the variation of \textit{normalized} $K$ values for a few randomly chosen filters indexes across layers and observe that it varies across layers, trying to match the distribution of weights at each layer. Each filter has its own set of weights, accounting for the differences in variation of $K$ in each case, as shown in Figure \ref{fig:kacrosslayers}.

\section{Conclusion}
We have proposed an optimal binary representation for network layer-weights that takes into account the distribution of weights, unlike previous distribution-agnostic approaches. We showed how this representation could be computed efficiently in $n.logn$ time using dynamic programming, thus enabling efficient training on larger datasets. We applied our technique on various datasets and noted significant accuracy improvements over other full-binarization approaches. We believe that this work provides a new perspective on network binarization, and that future work can gain significantly from distribution-aware explorations. 
{\small
\bibliographystyle{ieee}
\bibliography{egbib}
}

\section{Appendix}
\newcommand{\eqn}[1]{\begin{equation}\begin{split} #1 \end{split}\end{equation}}
\newcommand{\x}{{\bf x}}
\newcommand{\y}{{\bf y}}
\newcommand{\z}{{\bf z}}
\newcommand{\s}{\sigma}

\appendix

\section{Introduction}
The supplementary material consists of the following:
\begin{enumerate}
\item Proof for Theorem 2 (Optimal representation of $\widetilde{\mathbf{W}}$) which provides us with the optimal values of $\alpha$, $\beta$ and $\mathbf{e}$ to represent $\mathbf{W}$.
\item Proof for Theorem 4 (Gradient derivation) which allows us to perform back-propagation through the network.
\item Proof for Theorem 1 (Expressibility proof)
\item Experimental details
\end{enumerate}

\section{Optimal representation of $\widetilde{\mathbf{W}}$}
\newtheorem{name2}{Theorem 2.}
\begin{name2}
\label{approx}
The optimal binary weight $\widetilde{\mathbf{W}}$ which minimizes the error function $\mathbf{J}$ is
$$\widetilde{\mathbf{W}} = \alpha \mathbf{e} + \beta \mathbf{(1-e)} \\$$
where $\; \alpha =\frac{\mathbf{W}^{T}\mathbf{e}}{K} \;$, $\; \beta = \frac{\mathbf{W}^{T}\mathbf{(1-e)}}{n-K} \;$ and the optimal $\mathbf{e}^\ast$ is $\;  \mathbf{e}^\ast  = \underset{\mathbf{e},K}{\mathrm{argmax}} (\frac{ \parallel \mathbf{W}^T\mathbf{e} \parallel^{2}}{K} + \frac{\parallel \mathbf{W}^{T}\mathbf{(1-e)}\parallel^{2}}{n-K})$
\end{name2}

\begin{proof}
The approximated weight vector $\widetilde{\mathbf{W}}=[\alpha \alpha \ldots \beta  \alpha  \ldots \beta \beta$] can be decomposed as:
$$[\alpha \alpha \ldots \beta  \alpha  \ldots \beta \beta] = \alpha \cdot [ 1 1 \ldots 0  1  \ldots 0 0 ] + \beta \cdot [ 0 0 \ldots 1  0  \ldots 1 1 ]$$ 
where without loss of generality, $\mathbf{e} \in \{0,1\}^n, \mathbf{e}^T\mathbf{e} >  0, \mathbf{e} \in \{0,1\}^n, (\mathbf{1-e})^T(\mathbf{1-e}) >  0$ and $\alpha,\beta \in \mathbb{R}$. This is because the trivial case where $\mathbf{e} = \mathbf{0}$ or $\mathbf{e} = \mathbf{1}$ is covered by substituting  $\alpha = \beta$ instead and the equation is independent of $\mathbf{e}$.
We have to find the values of $\alpha$ and $\beta$ which would be the best approximation of this vector. \\
Let us define the error function $\mathbf{J} = \mathbf{W} - (\alpha \cdot \mathbf{e} + \beta \cdot (\mathbf{1-e}))$.
We have to minimize $\parallel \mathbf{J} \parallel^2 = \mathbf{E}$, where:
\begin{dmath}
\mathbf{E} = (\mathbf{W} - (\alpha \cdot \mathbf{e} + \beta \cdot (\mathbf{1-e})))^T(\mathbf{W} - (\alpha \cdot \mathbf{e} + \beta \cdot (\mathbf{1-e}))) \end{dmath}
\begin{dmath}
\mathbf{E} = \mathbf{W}^{T}\mathbf{W} + \alpha^2 \cdot \mathbf{e}^T\mathbf{e} + \beta^2\mathbf{(1-e)}^T \mathbf{(1-e)} \\ - 2\alpha \cdot \mathbf{W}^T\mathbf{e} - 2\beta \cdot \mathbf{W}^T(\mathbf{1-e}) + 2\alpha \beta \mathbf{e}^{T}\mathbf{(1-e)}
\end{dmath}
where $\mathbf{e}^T\mathbf{e} = K$, then $(\mathbf{1-e})^T(\mathbf{1-e}) = n-K$ and $\mathbf{e}^{T}\mathbf{(1-e)} = 0$. Substituting these in, we get
\begin{equation}\mathbf{E} =  \mathbf{W}^{T}\mathbf{W} + \alpha^2 K + \beta^2 (n-K) -\\ 2\alpha \cdot \mathbf{W}^T\mathbf{e} - 2\beta \cdot \mathbf{W}^T(\mathbf{1-e}) \label{wformulation} 
\end{equation}
We minimize this equation with respect to $\alpha$ and $\beta$ giving us:

\begin{equation}\frac{\partial \mathbf{E}}{\partial \alpha} = 0 , \frac{\partial \mathbf{E}}{\partial \beta} = 0 \end{equation}
Solving the above, we get the equations:
$$ \frac{\partial \mathbf{E}}{\partial \alpha} = 2 \alpha K - 2 \cdot \mathbf{W}^T\mathbf{e} = 0 $$ $$ \frac{\partial \mathbf{E}}{\partial \beta} = 2 \beta (n-K) - 2 \cdot \mathbf{W}^T(\mathbf{1-e}) = 0  $$
We can get the values of $\alpha$ and $\beta$ from the above equations.
$$ \alpha = \frac{\mathbf{W}^T\mathbf{e}}{K} , \beta = \frac{\mathbf{W}^T(\mathbf{1-e})}{(n-K)} $$ 
Then substituting the values of $\alpha$ and $\beta$ in equation \ref{wformulation}, we get
\begin{dmath} 
\mathbf{E} = \parallel \mathbf{W} ||^2 + \frac{ \parallel \mathbf{W}^T\mathbf{e} \parallel^{2}}{K} + \frac{\parallel \mathbf{W}^{T}\mathbf{(1-e)}\parallel^{2}}{n-K} - 2\frac{\parallel \mathbf{W}^{T}\mathbf{e} \parallel^{2}}{K} - 2\frac{\parallel \mathbf{W}^{T}\mathbf{(1-e)}\parallel^{2}}{n-K} 
\end{dmath}
\begin{dmath}
\mathbf{E} = \parallel \mathbf{W} ||^2 - (\frac{ \parallel \mathbf{W}^T\mathbf{e} \parallel^{2}}{K} + \frac{\parallel \mathbf{W}^{T}\mathbf{(1-e)}\parallel^{2}}{n-K})
\end{dmath}
In the above equation, we want to minimize $\mathbf{E}$. Since $\mathbf{W}$ is a given value, we need maximize the second term to minimize the expression. For a given $K$, $\mathbf{e_K} = sgn(\mathbf{T_k})$ where $\mathbf{T_k} = topk(\mathbf{W},K)$.
Here, $topk(\mathbf{W},K)$ represents the top $K$ values of $\mathbf{W}$ corresponding to either the largest positive $K$ values or the largest negative $K$ values, which remain as is whereas the rest are converted to zeros.

$$ \mathbf{e}^\star  = \underset{\mathbf{e,K}}{\mathrm{argmax}} (\frac{ \parallel \mathbf{W}^T\mathbf{e} \parallel^{2}}{K} + \frac{\parallel \mathbf{W}^{T}\mathbf{(1-e)}\parallel^{2}}{n-K}) $$

Selecting the $topk(\mathbf{W},K)$ would be optimal since $||\mathbf{W}^{T}\mathbf{e}||$ and $||\mathbf{W}^{T}\mathbf{(1-e)}||$ are both maximized on selecting either the largest $K$ positive values or the largest $K$ negative values. Hence, this allows us to select the optimal $\mathbf{e}$ given a $K$.

With this, we obtain the optimal $\mathbf{e}$.
\end{proof}

\section{Gradient derivation}

$$W \approx \widetilde{\mathbf{W}} = \alpha \mathbf{e} + \beta \mathbf{(1-e)} \\$$
$$where\; \alpha =\frac{\mathbf{W}^{T}\mathbf{e}}{K} \; and \; \beta = \frac{\mathbf{W}^{T}\mathbf{(1-e)}}{n-K} \\$$
Let $\mathbf{T_k} = topk(\mathbf{W}, K)$, and $\widetilde{\mathbf{W}_{1}} = \alpha \mathbf{e}$, and $\widetilde{\mathbf{W}_{2}} = \beta \mathbf{(1-e)}$. \\
Considering $\alpha$, on substituting $e = sgn(T_k)$. \\
$$\alpha = \frac{\mathbf{W}^{T}\mathbf{e}}{K} \\$$
$$\therefore \alpha = \frac{\mathbf{W}^{T}sgn(\mathbf{T_k})}{K}$$
Hence, we have $\alpha = \frac{\mathbf{W}^{T}sgn(\mathbf{T_k})}{K}$ and similarly $\beta = \frac{\mathbf{W}^{T}(1-sgn(\mathbf{T_k}))}{n-K}$. Putting these back in $\widetilde{\mathbf{W}}$, we have, \\
\begin{dmath}
\therefore \widetilde{\mathbf{W}} = \frac{\mathbf{W}^{T}sgn(\mathbf{T_k})}{K}\circ sgn\mathbf{(T_k)} + \frac{\mathbf{W}^{T}(1-sgn(\mathbf{T_k}))}{n-K}\circ (1-sgn\mathbf{(T_k)})
\end{dmath}
Now, we compute the derivatives of $\alpha$ and $\beta$ with respect to $\mathbf{W}$,
$$\frac{d\alpha }{d\mathbf{W}} = \frac{d(\mathbf{W}^{T}sgn(\mathbf{T_k}))}{d\mathbf{W}}.\frac{1}{K} \\$$
$$\frac{d\alpha }{d\mathbf{W}} = \frac{d(\mathbf{T_k}^{T}sgn(\mathbf{T_k}))}{d\mathbf{W}}.\frac{1}{K} \\$$
\begin{dmath}
\frac{d\alpha }{d\mathbf{W}} = \frac{d(||\mathbf{T_k}||_{l1})}{d\mathbf{W}}.\frac{1}{K}=\frac{sgn(\mathbf{T_k})}{K}
\end{dmath}
Similarly, \\
\begin{dmath}
\frac{d\beta}{d\mathbf{W}} = \frac{d(||\mathbf{W}-\mathbf{T_k}||_{l1})}{d\mathbf{W}}.\frac{1}{n-K}=\frac{sgn(\mathbf{W}-\mathbf{T_k})}{n-K}
\end{dmath}
Now, $\widetilde{\mathbf{W}_{1}} = \alpha \mathbf{e}$ therefore,
$$\frac{d\widetilde{\mathbf{W}_{1}}}{d\mathbf{W}} = e \frac{d\alpha}{d\mathbf{W}} + \alpha \frac{d\mathbf{e}}{d\mathbf{W}} $$
$$\therefore \frac{d\widetilde{\mathbf{W}_{1}}}{d\mathbf{W}} = \frac{sgn(\mathbf{T_k})}{K}\circ sgn(\mathbf{T_k}) + \alpha.STE(\mathbf{T_k})$$
With this, we end up at the final equation for $\widetilde{\mathbf{G_1}} = \frac{d\widetilde{\mathbf{W}_{1}}}{d\mathbf{W}}$ as mentioned in the paper,
\begin{dmath}
\therefore \widetilde{\mathbf{G_1}} = \frac{sgn(\mathbf{T_k})}{K} \circ sgn(\mathbf{T_k}) + \frac{||\mathbf{T_k}||_{l1}}{K}STE(\mathbf{T_k})
\end{dmath}
Considering the second term $\widetilde{\mathbf{W}_2}$, we have,
$$\frac{d\widetilde{\mathbf{W}_{2}}}{d\mathbf{W}} = \mathbf{(1-e)} \frac{d\beta}{d\mathbf{W}} + \beta \frac{d\mathbf{(1-e)}}{d\mathbf{W}}$$
$$\therefore \frac{d\widetilde{\mathbf{W}_{2}}}{d\mathbf{W}} = \frac{sgn(\mathbf{W}-\mathbf{T_k})}{n-K} \circ (1-sgn(\mathbf{T_k})) + \beta.STE(\mathbf{W}-\mathbf{T_k})$$
This provides us $\widetilde{\mathbf{G_2}} = \frac{d\widetilde{\mathbf{W}_{2}}}{d\mathbf{W}}$ as mentioned in the paper,
\begin{dmath}
\widetilde{\mathbf{G_2}} = \frac{sgn(\mathbf{W}-\mathbf{T_k})}{n-K} \circ (1-sgn(\mathbf{T_k})) + \frac{||\mathbf{W}-\mathbf{T_k}||_{l1}}{n-K}.STE(\mathbf{W}-\mathbf{T_k})
\end{dmath}
Together, we arrive at our final gradient $\widetilde{\mathbf{G}} = \frac{d\widetilde{\mathbf{W}}}{d\mathbf{W}}$,
\begin{dmath}
\widetilde{\mathbf{G}} = \widetilde{\mathbf{G_1}} + \widetilde{\mathbf{G_2}}
\end{dmath}

\section{Binary Networks as Approximators}
We define $m_{k}$ as the number of neurons required to approximate a polynomial of $n$ terms, given the network has a depth of $k$. We show that this number is bounded in terms of $n$ and $k$.
\begin{theorem}
For $p(\x)$ equal to the product $x_1x_2\cdots x_n$, and for any $\s$ with all nonzero Taylor coefficients, we have:
\begin{equation}
m_k(p, \s) = \mathcal{O}\left(n^{(k-1)/k}\cdot 2^{n^{1/k}}\right).\label{eqn:constantlayers}
\end{equation}
\label{thm:constantlayers}
\end{theorem}
\begin{proof}
We construct a binary network in which groups of the $n$ inputs are recursively multiplied.  The $n$ inputs are first divided into groups of size $b_1$, and each group is multiplied in the first hidden layer using $2^{b_1}$ binary neurons (as described in \cite{lin2017does}).  Thus, the first hidden layer includes a total of $2^{b_1}n/b_1$ binary neurons. This gives us $n/b_1$ values to multiply, which are in turn divided into groups of size $b_2$.  Each group is multiplied in the second hidden layer using $2^{b_2}$ neurons.  Thus, the second hidden layer includes a total of $2^{b_2}n/(b_1b_2)$ binary neurons.

We continue in this fashion for $b_1, b_2, \ldots, b_k$ such that $b_1b_2\cdots b_k = n$, giving us one neuron which is the product of all of our inputs.  By considering the total number of binary neurons used, we conclude
\begin{equation}
m_k(p, \s) \le \sum_{i = 1}^k \frac{n}{\prod_{j = 1}^i b_j} 2^{b_i} = \sum_{i = 1}^k \left(\prod_{j = {i + 1}}^k b_j\right) 2^{b_i}.
\label{eqn:branching}
\end{equation}
Setting $b_i = n^{1/k}$, for each $i$, gives us the desired bound (\ref{eqn:constantlayers}).
\end{proof}

\section{Expressibility of Binary Networks}
A binary neural network (a network with weights having only two possible values, such as $+1$ and $-1$) with a single hidden layer of $m$ binary-valued neurons that approximates a product gate for $n$ inputs can be formally written as a choice of constants $a_{ij}$ and $w_j$ satisfying
\begin{equation}
\sum_{j=1}^m w_j \sigma\left(\sum_{i=1}^n a_{ij} x_i\right) \approx \prod_{i=1}^n x_i. \label{goal}
\end{equation}

\cite{lin2017does} shows that $2^n$ neurons are sufficient to approximate a product gate with $n$ inputs - each of these weights are assigned, in the proof, a value of $+1$ or $-1$ before normalization, and all coefficients $a_{ij}$ also have $+1$/$-1$ values. This essentially makes it a binary network. Weight normalization introduces a scaling constant of sorts, $\frac{1}{2^{n}n!\sigma_{n}}$, which would translate to $\alpha$ in our representation, with its negative denoting $\beta$. \\
The above shows how binary networks are expressive enough to approximate real-valued networks, without the need for higher bit quantization.
\section{Experimental details}
We used the Adam optimizer for all the models with a maximum learning rate of 0.002 and a minimum learning rate of 0.00005 with a decay factor of 2. All networks are trained from scratch. Weights of all layers except the first were binarized throughout our experiments. Our FBin layer is structured the same as the XNOR-Net. We performed our experiments using a cluster of GeForce GTX 1080 Tis using PyTorch v0.2.

Note: The above proofs for expressibility power have been borrowed from \cite{lin2017does}.

\end{document}